\documentclass[final]{cvpr}
\usepackage{diagbox}
\usepackage{times}
\usepackage{epsfig}
\usepackage{graphicx}
\usepackage{amsmath}
\usepackage{amssymb}
\usepackage{algorithm}
\usepackage{algorithmicx}
\usepackage{algpseudocode}
\usepackage{booktabs}
\usepackage{array}
\usepackage{colortbl}
\usepackage{overpic}
\usepackage[table]{xcolor}
\usepackage{multirow}
\usepackage{bbding}
\usepackage{pythonhighlight}
\usepackage{algorithm}
\usepackage{algorithmicx}
\usepackage{algpseudocode}  
\usepackage{bm}
\usepackage{pifont}
\def\ie{\emph{i.e.,~}}
\def\eg{\emph{e.g.,~}}

\newcommand{\minisection}[1]{\vspace{0.0in} \noindent {\bf #1}}

\definecolor{mygreen}{RGB}{0,150,0}
\definecolor{myred}{RGB}{200,0,0}
\usepackage[pagebackref,breaklinks,colorlinks]{hyperref}

\def\cvprPaperID{6586} % *** Enter the CVPR Paper ID here
\def\confYear{CVPR 2023}
\graphicspath{{../figures/}}

\newlength\savedwidth
\newcommand{\whline}[1]{\noalign{\global\savedwidth\arrayrulewidth \global\arrayrulewidth #1}%
                   \hline \noalign{\global\arrayrulewidth\savedwidth}}

\begin{document}

\title{SLAN: Self-Locator Aided Network for Cross-Modal Understanding}

\author{ Jiang-Tian Zhai$^{1}$\thanks{Indicates equal contributions.} \thanks{Work done when interning at Tencent Youtu Lab.}
    \quad  \quad Qi Zhang$^{2}$\footnotemark[1] \quad \quad 
    Tong Wu$^2$ \quad \quad Xing-Yu Chen$^2$ \quad \quad Jiang-Jiang Liu$^1$\footnotemark[2]\\ Bo Ren$^2$  \quad \quad Ming-Ming Cheng$^1$ \\
    CS, Nankai University$^1$ \quad \quad Tencent Youtu Lab$^2$ \\
    {\tt\small scok30@sina.com}, {\tt\small \{labyrinth7x,clinene0322,j04.liu\}@gmail.com}\\
    {\tt\small \{townswu,timren\}@tencent.com}, {\tt\small cmm@nankai.edu.cn}
}
\maketitle
\thispagestyle{empty}

%%%%%%%%% ABSTRACT
\begin{abstract}
Learning fine-grained interplay between vision and language allows to a more accurate understanding for Vision-Language tasks. However, it remains challenging to extract key image regions according to the texts for semantic alignments. Most existing works are either limited by text-agnostic and redundant regions obtained with the frozen detectors, or failing to scale further due to its heavy reliance on scarce grounding (gold) data to pre-train detectors. To solve these problems, we propose Self-Locator Aided Network (SLAN) for cross-modal understanding tasks without any extra gold data. SLAN consists of a region filter and a region adaptor to localize regions of interest conditioned on different texts. By aggregating cross-modal information, the region filter selects key regions and the region adaptor updates their coordinates with text guidance. With detailed region-word alignments, SLAN can be easily generalized to many downstream tasks. It achieves fairly competitive results on five cross-modal understanding tasks (\eg 85.7\% and 69.2\% on COCO image-to-text and text-to-image retrieval, surpassing previous SOTA methods). SLAN also demonstrates strong zero-shot and fine-tuned transferability to two localization tasks. 

\end{abstract}

%%%%%%%%% BODY TEXT
\section{Introduction} \label{sec:introduction}
%background: few lines
% Vision-Language problem includes a wide range of tasks, including image-text retrieval, image caption, visual question answer, visual grounding, \etc. The rapid development for vision-language problem boosts many downstream applications, \eg multi-modal search engines, video caption generators and recommender systems.  One basic requirement of these tasks is to learn joint representations for both  modalities.   
Recent years have witnessed growing interest in exploring relationships between vision and language modalities. A wide range of applications have been boosted by its rapid development, such as multi-modal search engines~\cite{chen2017amc,elizalde2019cross,cao2017collective} and recommender systems~\cite{sun2020multi,wang2019learning,chen2020imram}. It motivates researchers to find semantic correspondence between two modalities and bridging their visual-semantic discrepancy. Some earlier works~\cite{radford2021learning,jia2021scaling,li2021align,gan2020large} focused on learning joint embeddings for these two modalities, while more recent ones~\cite{kamath2021mdetr,li2022grounded,zhang2021vinvl} have turned to considering latent vision-language alignments at the level of regions and words.

%recent works and their problems

\newcommand{\addff}[1]{\includegraphics[width=0.48\linewidth]{fig/fig1/#1.png}}
\begin{figure}
	\centering
	\small
	\renewcommand{\arraystretch}{1.1}
  \setlength\tabcolsep{1.2mm}
	
\begin{tabular}{cc}

\addff{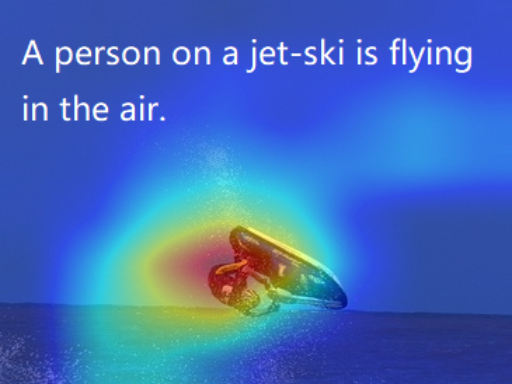}&\addff{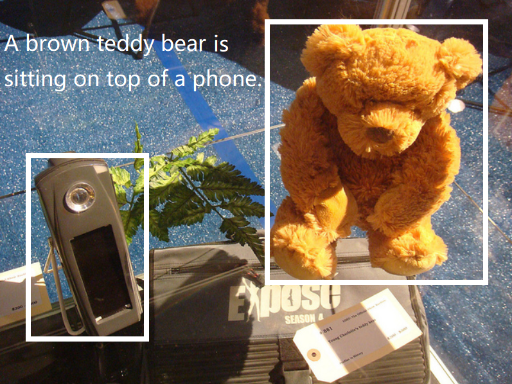}\\
(a) Image-text Retrieval&(b) Image Caption\\

\addff{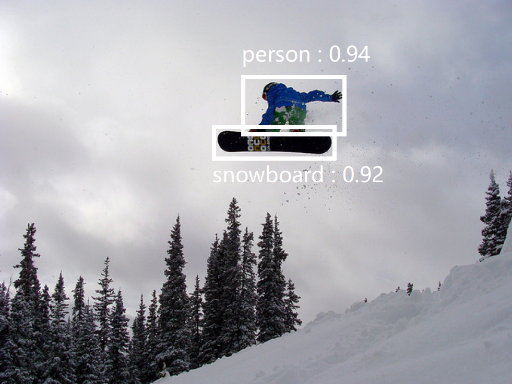}&\addff{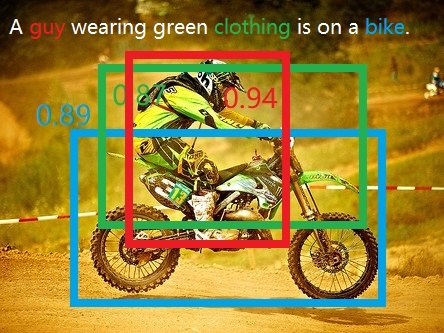}\\
(c) Object Detection&(d) Phrase Grounding\\
    \end{tabular}
    \caption{Visualization on four different tasks.
    We visualize the activation map for text-to-image retrieval task in (a). As for the caption task in (b), we visualize regions selected by our model.
    %, which are class-agnostic since there is no text input. 
    Besides cross-modal understanding task, SLAN can transfer to localization tasks, shown in (c) and (d), and we list the confidence score for each region.}
	\label{fig:fig1}
\end{figure}

% \zjt{frozen detector}
% Fine-grained cross-modal alignments help model learn more accurate understanding for downstream tasks, some works use the object detector as a crucial component for many cross-modal understanding systems. They treat it as a black-box to extract object-centric regions for alignments with word embeddings, which keep it frozen during the cross-modal learning. This prevents the extracted key regions to adapt with downstream cross-modal tasks. In comparison, VinVL applies a pre-trained object detector with more than 2000 classes and attributes to enrich local visual representations. However, compared with the free-form text from a larger-scale cross-modal dataset, the expanded label set still limits the perceptive capability of the object detector for cross-modal understanding. 
In order to achieve fine-grained cross-modal alignments, some works~\cite{lee2018stacked,kuo2022beyond,li2020oscar} use object detectors  to extract key regions in images. Treated as black boxes, the detectors only support for fixed vocabulary object detection. Meanwhile, the extracted regions cannot adapt to different text information due to the freezing parameters of the detectors. To alleviates the problem, VinVL~\cite{zhang2021vinvl}  applies a pre-trained object detector with more than $2000$ classes and attributes to enrich local visual representations. However, the extended label set still limits the perceptive capability of object detectors for cross-modal understanding compared to free-form text from real-world scenes.

Recently, more works have attempted to apply learnable region locators for cross-modal tasks, which extract regions of interest conditioned on different texts. Unlike previous methods using frozen object detectors,  MDETR~\cite{kamath2021mdetr} builds an end-to-end framework on datasets with region-to-word annotations.  GLIP~\cite{li2022grounded} directly proposes grounded language-image pre-training for learning object-level,  language-aware, and semantic-rich visual representation. These methods demonstrate their effectiveness in cross-modal reasoning by introducing trainable locators. However, in order to supervise the training of locators, these methods require a certain amount of region-to-word grounding annotations (gold data), which are based on burdensome and expensive annotation efforts. It limits their applications on existing larger scale of cross-modal datasets which have abundant but coarse-grained image and text pairs.

To address the problems above, we propose Self-Locator Aided Network (SLAN) for cross-modal understanding. The designed self-locator is capable of accurately locating regions of interest based on different texts. Specifically, the self-locator consists of a region filter to select important regions and a region adaptor to update coordinates of regions with text guidance. By incorporating the self-locator into our framework, SLAN performs context-aware region extraction and cross-modal feature fusion. Moreover, SLAN is trained solely on datasets with paired images and texts, making it scalable to larger pre-training settings for further performance improvements. With fine-grained region-word alignments, SLAN has a more detailed understanding of interactions in vision and language modalities.

%conclude
To sum up, our contributions have four aspects:
%\begin{itemize}
%    \item We introduce a framework termed SLAN incorporating a self-locator to propose regions matched with text. %This enables dynamic word-to-region alignments for cross-modal understanding tasks. 
%    \item  The framework does not use gold data for supervision, and can be applied to the large-scale pre-training with datasets having paired images and texts.
%    \item  Our proposed SLAN is able to locate text-aware key regions and can naturally transfer to some localization tasks, object detection and phrase grounding.
%    \item Experiments on five cross-modal understanding and two localization tasks demonstrate the effectiveness of our method. For example, SLAN achieves state-of-the-art performance on COCO image-text retrieval. 
%\end{itemize}
\begin{itemize}
    \item We propose a framework termed SLAN to capture fine-grained interplay between vision and language modalities. A self-locator is introduced to perform text-guided region adaptation, enabling dynamic region-word alignments for cross-modal understanding tasks.
    \item We demonstrate that SLAN can be easily applied to large-scale pre-training on cross-modal datasets, because it is free from training with gold data.
    \item We naturally generalize SCAN to some localization tasks, such as object detection and phrase grounding, due to its ability to locate key regions in images.
    \item Experiments on five cross-modal understanding and two localization tasks demonstrate the effectiveness of our method. For example, SLAN achieves state-of-the-art performance on COCO image-text retrieval. 
\end{itemize}

\section{Related Work}
\subsection{Vision-language Task}
\begin{figure*}[!t]
    \centering
    \includegraphics[scale=0.215]{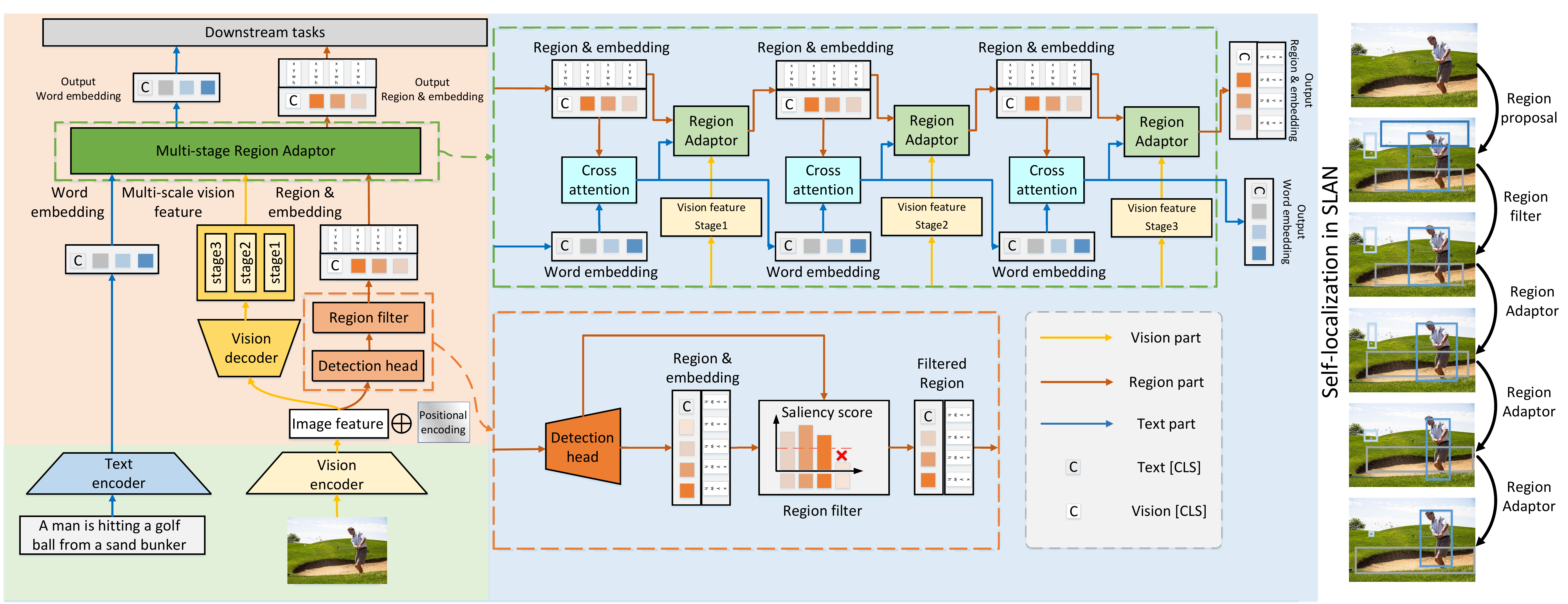}
	\caption{Overall framework of our proposed SLAN, consisting of two unimodal encoders for embedding images and text and a self-locator for fine-grained region-word alignments.  The self-locator performs saliency prediction with a region filter and progressive regression  with a region adaptor. The learned vision and language features with cross-modal awareness are used for downstream tasks.
	}\label{fig:main}
\end{figure*}
% Vision-language tasks aim to align visual with language representation and applies this knowledge to various downstream tasks. Some custom methods design loss and network structure to learn the alignment. DeViSE~\cite{frome2013devise} first introduces a linear layer to unify image and text embedding. Two-Branch Neural Networks~\cite{wang2018learning} learns cross-modal features and applies contrastive loss. Instance loss is introduced in ~\cite{zheng2020dual} to train the dual encoder. Other methods use some prior tool or knowledge to align these cross-modal features. For example, SGG~\cite{xu2017scene} considers the relation between these image regions and compares it with the text structure. ViSTA applies OCR to extract text embeddings from images to benefit from richer cross-modal labels. 

Many efforts have been made to explore relationships between visual and textual modalities, then apply the knowledge to various downstream multi-modal tasks. Some previous methods propose loss functions and network structure to learn semantic vision-language alignments. DeViSE~\cite{frome2013devise} first introduces a linear layer to unify image and text embeddings. TBNN\cite{wang2018learning} learns cross-modal features and applies contrastive losses. The instance loss is introduced in ~\cite{zheng2020dual} to train the dual encoder. Other methods introduce
some prior tools or knowledge to assist the image-text matching analysis. For example, SGG~\cite{xu2017scene}  considers the internal relations between visual regions and compares them with the text structure. ViSTA~\cite{cheng2022vista} applies OCR~\cite{biten2019scene} to extract text from images for richer cross-modal labels.

In recent years, a prevailing direction applies cross-modal pre-training on some larger datasets. CLIP~\cite{radford2021learning} pre-trains with 400M image-text pairs collected from the web, building global relations between images and texts. BLIP~\cite{li2022blip} benefits from large-scale web data by filtering out noisy ones, and performs vision-language understanding and generation tasks. Beit-3~\cite{wang2022image} adopts mask-then-predict self-supervised training on large-scale monomodal data (\ie images, and texts) and multi-modal data (\ie image-text pairs) to learn internal cross-modal dependencies.

Aside from these attempts, exploring local relations between words in text and objects in the image works efficiently in cross-modal pre-training. It helps localize more accurate objects according to corresponding words, and provides cues for downstream tasks.

% GLIP~\cite{li2022grounded} proposes grounded vision-language pre-training on fine-grained annotations to learn accurate cross-modal alignment.

% Our method inherits this benefit without the requirement on fine-grained annotations.

\subsection{Localization for Vision-language Task}
% Localizing image regions and words in sentence helps model learn local alignment. 
There are two kinds of methods whose differences are whether the detection module are frozen or trained to adapt to cross-modal tasks.
% and the scale of fine-grained labeled data for localization. 
The first kind introduces a frozen object detector to extract detailed visual representations. SCAN uses Faster R-CNN pre-trained on Visual Genomes to choose key regions for matching with word embeddings. 
Instead of using a traditional object detector trained with relatively few classes and data, some later works (e.g., VinVL~\cite{zhang2021vinvl}, Oscar~\cite{li2020oscar}) increase the number of detection labels and introduce some attribute information to complement previous visual concepts. 
% This makes the extracted visual representation more adaptive to cross-modal alignments. 

The other kind relies on fine-grained annotations of the cross-modal dataset to  perform pre-training. MDETR~\cite{kamath2021mdetr} introduces a modulated detector with multi-modal datasets, which have precise alignments between phrases in text and objects in the images. GLIP~\cite{li2022grounded} applies grounded pre-training to learn object-level, language-aware, and semantic-rich visual representations. These methods link object detection with phrase grounding (region-word alignments), enabling models to learn from richer semantic knowledge from both modalities. However, these methods require cross-modal data with fine-grained annotations, limiting their application on larger-scale pre-training settings. 
% Therefore we propose our ... to learn local alignment with coarse image-text pairs. 

\section{SLAN}
We will introduce our proposed SLAN in this section. As shown in Figure~\ref{fig:main}, our framework consists of three components, two unimodal encoders and a self-locator. We first briefly introduce two unimodel encoders.
Then we introduce the detailed structures of SLAN which adaptively select informative regions with text guidance. Finally, we list our pre-training objectives. 
%without supervision from gold data. 
% \subsection{Overall Architecture}
% Our main framework has four components as shown in Figure~\ref{fig:main}: two unimodal encoders, a vision decoder, and a self-locator. 
% vision encoder, a text encoder, a vision decoder, and the self-locator. 
%and the following module for downstream tasks. 

% \minisection{Unimodal Encoder.} 
\subsection{Unimodal Encoding}
% We introduce the vision and text encoder for extracting visual and textual representations.
%as illustrated in Figure~\ref{fig:main}. 
% We use BERT as our text encoder, where a [CLS] token is added to summarize the whole sentence. We apply the global average pooling on the outputed vision feature map, and treat it as the vision [CLS] token. The [CLS] tokens of two modalities are used for downstream tasks and pre-training. 
We introduce the vision and text encoder to learn visual and textual representations with ${D}$ dimensions. For text feature extraction, we use BERT~\cite{kenton2019bert} as our text encoder, encoding words into a shared semantic space. A text [CLS] token is added to word embeddings to summarize the whole sentence. For image feature extraction, we encode images and obtain the vision feature map $V$ with high-level semantics.

\subsection{Self-locator for Cross-modal Understanding}\label{ss:mod}
Since fine-grained region-word alignments are important for cross-modal relation exploration, our self-locator is built on DETR~\cite{carion2020end} to output original regions. Region embeddings are extracted from the feature map $V$. A vision [CLS] token is then obtained from global average pooling of region embeddings.
% The [CLS] tokens of two modalities are used for downstream tasks and pre-training.

Different from most traditional object detection tasks that use the pre-defined label set, cross-modal tasks usually have a wider vocabulary and free-form textual expressions. Therefore, our self-locator is adapted to introduce a region filter for region saliency prediction and a region adaptor for progressive region regression for cross-modal tasks. By replacing fixed vocabulary prediction with region saliency prediction, our self-locator assigns each region a saliency score to estimate the probability that the region is useful for the alignment process. For traditional detection settings, the regression targets are annotated region coordinates. Since there is no grounding (gold) annotations in our setting, we propose progressive region regression supervised by a weighted summation of updated regions.

\subsubsection{Vision Decoder: Pyramid Feature Extraction}

Our proposed self-locator is designed for regression in a coarse-to-fine manner, requiring visual features of multi-scale and pyramid hierarchy. Considering these characteristics, we adopt a vision decoder after the global visual feature to extract multi-scale feature maps. Let $F_i ~(i \in {1,2,...,L})$ denotes the $i$-th level of decoder features separately, where $L$ is the number of layers of the self-locators. Then $F_i$ is fed to the $i$-th level of self-locator regression.

\subsubsection{Region Filter: Region Saliency Prediction}
%Original detection modules (e.g., DETR) output a relatively large number of region proposals (e.g., 100). 
% As for cross-modal tasks, the learned visual features should align with text ones.
% Considering the caption of an image, the number of mentioned object in this sentence is actually less than 100. If we directly select all detected regions, there will be some redundant regions involved for learning cross-modal alignment. This leads to unnecessary computational cost and may introduce some meaningless region-to-word pairs for the model to learn. To solve this problem, we set a parameter $T$ to control the maximum number of regions selected after the region filter. The strategy is as follows: (a) Normalize all saliency scores $S_i$ of these regions. After this process the maximum score is 1. (b) Sort these regions according to their saliency scores $S_i$ in descending order. (c) Select regions with saliency scores higher than a threshold $h$ ($S_i>h$). (d) If the number of selected regions is more than $T$, the filter picks the first $T$ regions. We then weight region embeddings by the scores. 
When describing images, people usually focus on limited salient regions in the images. However, original detection models (e.g., DETR) output a relatively large number of region proposals (e.g., 100) for images, which are mostly redundant for text descriptions. If we directly select all detected regions, it will lead to unnecessary computational cost and might cause the model to learn some meaningless region-to-word pairs. The strategy to control the maximum number of selected regions. The strategy is as follows: (a) Normalize all saliency scores of these regions. After this process, the scores are represented as $S=\{S_1, ...,S_k\}, S_i \in \mathbb{R}$, with the maximum value of 1. (b) Sort these regions in descending order according to their saliency scores. (c) Select regions with saliency scores above the threshold $h$ ($S_i>h$). (d) If the number of selected regions is greater than  hyper-parameter $T$, the self-locator picks the first $T$ regions. Finally, we weight region embeddings by the scores.

\begin{figure}[!t]
    \centering
    \includegraphics[scale=0.35]{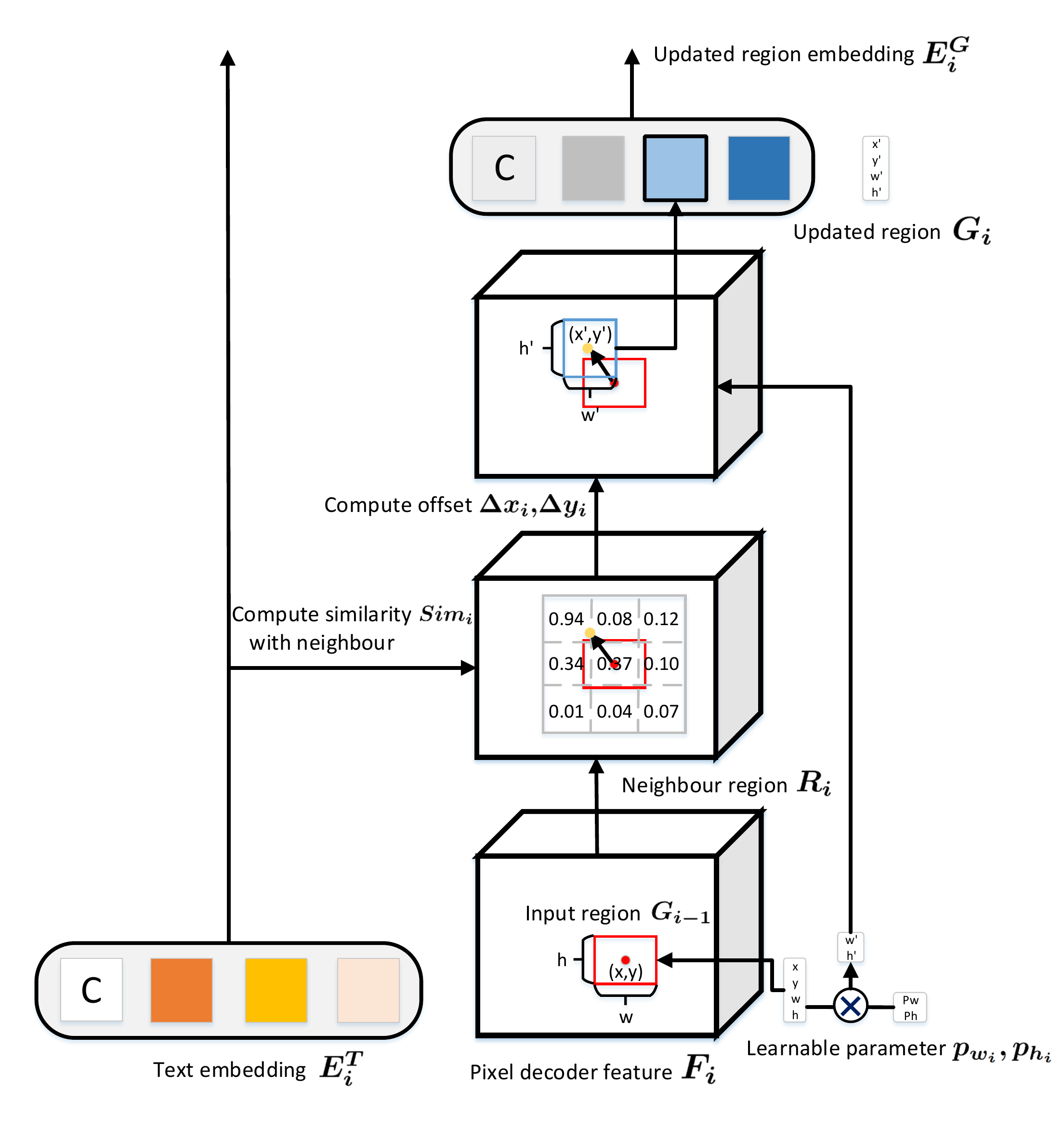}
	\caption{Illustration of one level of the region adaptor to update each region's coordinate with text guidance. We use the feature map from vision decoder to extract region embeddings and explore latent region-word alignments.
	}\label{fig:ada}
\end{figure}

\begin{algorithm}[h]
	\renewcommand{\algorithmicrequire}{\textbf{Input:}}
	\renewcommand{\algorithmicensure}{\textbf{Output:}}
	\caption{Self-localization} 
	\label{alg::ba} 
	\begin{algorithmic}[1] 
		\Require 
		Image $x$, region embeddings $E_0^G$, text embeddings $E_0^T$, pyramid feature map $F_i$, neighbour size $(NH_i, NW_i)$, total region regression layers $L$.
		\Ensure 
		Updated regions $G_{out}$, region supervision on detection head $\overline{G}$, visual token $v_{cls}$, textual token $t_{cls}$.
		
		\State $G_0, S, E_0^G$ ← Detection($x$) 
		\State $G_0, E_0^G$ ← Region Saliency Prediction ($G_0, S, E_0^G$)
		
		\For{$i\in$ $\{1,2,...,L\}$}
		
% 		\State $G'=\emptyset,E_G'=\emptyset$
		\State $E^G_i, E^T_i$ ← Cross attention($E^G_{i-1},E^T_{i-1}$)
		
% 		\For{$g \in G$}
% 		\State $x,y,w,h$ ← $g$
		
		\State $ R_i $ ← $(NH_i,NW_i)$
		\State $ E^{Ne}_i $ ← Neighbour Embedding($R_i,G_{i-1}$)
		\State $Sim_i$ ← Similarity($E^{Ne}_i$,$E^T_i$)
		\State $\Delta$$x_i$,$\Delta$$y_i$ ← $Offset$($Sim_i$)
		\State $p_{w_i}, p_{h_i}$ are learnable parameters.
		\State $G_i$ ← Update($G_{i-1}$,$\Delta$$x_i$,$\Delta$$y_i$,$p_{w_i}, p_{h_i}$)
		\State $E^G_i$ ← Embedding($G_i,f_i$)
% 		\State $ N_{gr} $ ← Split to Grids($N_{re}$,$K$)
% 		\State $E_n$ ← \{Embedding($r$,$f_i$)$| r\in$ $N_{gr}$\}
% 		\State $S_n$ ← $\emptyset$
% 		\For{$e_n \in E_n$}
% 		\State $S_n$ ← $S_n\cup$ max(\{sim($e_t,e_n$)$|e_t\in E_T$\})
% 		\EndFor
% 		\State $\Delta$$x$,$\Delta$$y$ ← $Offset$($S_n$)
% 		\State $g'$ ← $x$+$\Delta$$x$, $y$+$\Delta$$y$, $w\times s_w$, $h\times s_h$ 
% 		\State $G'$ ← $G'\cup g'$
% 		\State $E_G'$ ← $E_G'\cup$ Embedding($g',f_i$)
% 		\EndFor
% 		\State $G$ ← $G'$
% 		\State $E_G$ ← $E_G'$
	\EndFor
	\State $v_{cls}, t_{cls}$ ← ExtractCLS($E^G_{out}, E^T_{out}$)
	\State $G_{out}$ ← $G_{L}$, ~$\overline{G}$ ← $(\sum_{i=1}^{L}G_i)/L$
	\end{algorithmic} 
\end{algorithm}

\label{ss:ba}
\subsubsection{Region Adaptor: Progressive Region Regression}
% This module adapts selected regions under the guidance of text information. More specifically, it aims at adjusting the coordinates of proposed regions to align with words with the same semantics. 
This module aims at adjusting the coordinates of proposed regions to align with words with the same semantics. The difficulty comes from no annotated text-referenced regions as ground truths. We transform this problem into the $L$ cascaded coarse-to-fine progressive regression. We set $L$ to 3 in default. As shown in Figure~\ref{fig:ada}, the $i$-th level of the region regression receives three inputs: word embeddings $E_{i-1}^T \in \mathbb{R}^{N^T \times D}$, region embeddings $E_{i-1}^G \in \mathbb{R}^{N^G \times D}$ with their coordinates $G_i \in \mathbb{R}^{N^G \times 4}$, and a global decoder feature map $F_i \in \mathbb{R}^{H_i \times W_i \times D}$, where $N^T$ and $N^G$ denotes the number of words and selected regions respectively.

We describe the procedure for progressive region regression in Algorithm~\ref{alg::ba}. The cross-modality multi-head attention layers fuse region and word embeddings and model their interactions as follows: 
\begin{equation}
  \begin{aligned}
    \label{eqn:ca}
    Attn &= \frac{E_{i-1}^G E_{i-1}^{T\top}}{\sqrt{D}}\\
    E^G_{i} &= Softmax(Attn) E^T_{i-1}\\
    E^T_{i} &= Softmax(Attn^\top) E^G_{i-1} \\
  \end{aligned}
  ,
\end{equation}
where $D$ denotes the dimension of embeddings. With cross-modal semantics, the updated vision-aware word embeddings $E^T_{i}$ are able to guide region coordinate updates by searching for highly correlated regions around the original one. Specifically, the neighborhood of the region $g=(x,y,w,h)$ is defined as a region of size $(NH_i, NW_i)$ centered on it, where $NH_i$ and $NW_i$ are pre-defined parameters for the $i$-th level in region regression. And the neighborhood is split to $K \times K$ regions to compute region-word similarities, as shown in Figure~\ref{fig:ada}, where each region embedding is extracted with RoIAlign and average pooling from $F_i$. With different response scores to words, neighbor regions aggregate context information to the central one. The coordinate update for the central region is in the form of weighted summation of coordinates of neighbor center points as follows:

%In Algorithm~\ref{alg::ba}, Embedding(*,*) from line 11 and 19 is implemented with the RoIAlign and global average pooling. And offset computation (\ie$Offset(S_n)$) from line 16 is to compute the offset on the region center according to the similarity between grid regions and text embeddings:
%\begin{equation}
%  \begin{aligned}
%    \label{eqn:wo}
%    Offset(S_n) &= \sum_{i=0}^{K^2-1}s_i o_i\\
%    o_i &= (ow_i,oh_i)\\
%    ow_i &= (\lfloor\frac{i}{K}\rfloor-\lfloor\frac{K}{2}\rfloor)\times H_r\\
%    oh_i &= ((i\ mod K)-\lfloor\frac{K}{2}\rfloor)\times W_r
%  \end{aligned}
%\end{equation}

\begin{equation}
\begin{aligned}
    \label{eqn:wo}
    w' &= p_w w,  ~ ~ h' = p_h h, \\
    x' &= x + \Delta x, ~ ~ y' = y + \Delta y, \\
    \Delta x &= \sum_{j=0}^{K^2-1}s_j H_j (\lfloor\frac{j}{K}\rfloor-\lfloor\frac{K}{2}\rfloor), \\
      \Delta y &= \sum_{j=0}^{K^2-1}s_j W_j ((j~ mod~K) -\lfloor\frac{K}{2}\rfloor), \\
\end{aligned}
\end{equation}
where $\lfloor \cdot \rfloor$ is the round down operation, $p_w$ and $p_h$ are learnable parameters, and $s_j$ is the maximum similarity between the embedding of the $j$-th neighbor region and all word embeddings.

For each original region $g$, let $g_i$ denotes the updated version after the $i$-th layer of region regression. We take the average of them as the ground truth and apply the $L_1$ and GIoU regression loss:
\begin{equation}
  \begin{aligned}
    \label{eqn:re}
    \overline{g} &= \frac{\sum_{i=1}^{L}g_i}{L}\\
    \mathcal{L}_{reg}(g) &= \mathcal{L}_{L1}(g,\overline{g})+\mathcal{L}_{GIoU}(g,\overline{g})\\
  \end{aligned}
  .
\end{equation}
% The following process is applied to each box: (a) Define a neighbour region around $(X, Y)$, which is defined by the receptive field $(H_r, W_r)$. This receptive field is pre-defined for each level of adaptor. The upper left corner of this neighbour region is $(X-H_r/2, Y-W_r/2)$. (b) Split this neighbour region into $K \times K$ grids, as illustrated in Figure~\ref{fig:ada}. (c) For each grids $G_j$, use RoIAlign and average pooling to extract its embedding $GE_j \in \mathbb{R}^{D}$. (d) Compute pair-wise similarity between each word embedding and grid embedding, and save the results in a tensor $Sim \in \mathbb{R}^{N_T \times (K \times K)}$. (e) Select the max similarity score on each grids to get $G_{map} \in \mathbb{R}^{K \times K}$. (f) Each grid $i$ in Figure~\ref{fig:ada} has a relative offset $OX_i, OY_i$ if we treat $(X, Y)$ as the origin. $(dX, dY)$ is the weighted average of $OX_i, OY_i$ according to $G_{map}$. Detailed explanation is in Appendix. (g) Two learnable parameters $Pw$ and $Ph$ is used to multiply with $(W, H)$. Combined with the offset of $(X, Y)$, the box coordinate is updated to $(X', Y', W', H')$, where $X'=X+dX$, $Y'=Y+dY$, $W'=W \times Pw$ and $H'=H \times Ph$.
% (h) Use RoIAlign to collect new box embedding. 

\subsection{Pre-training Objectives with SLAN}
% Our SLAN receives image-text supervision from paired pre-training data, learning fine-grained region-to-word alignments from them. There are two losses for understanding and one loss for generation. 
Our SLAN pretrains on image-text pairs and learns fine-grained region-word alignments, supervised by three common losses.
 
\minisection{Image-Text Matching Loss (ITM)} predicts whether a given image-text pair is positive or not, which can be viewed as a binary classification problem. The visual and textual [CLS] tokens $(v_{cls}, t_{cls})$ are concatenated and sent to a linear layer $f_c$. The ITM loss is formalized as follows:
\begin{equation}
  \begin{aligned}
    \label{eqn:itm}
    \mathcal{L}_{itm} (I,T) &= H(f_c(cat(v_{cls}, t_{cls})),y_{v,t}),
  \end{aligned}
\end{equation}
where $y_{v,t}$ denotes the matching relation  (1 for matched and 0 for unmatched), and $H$ is the cross-entropy loss for classification. 
We directly select positive pairs from the dataset and build hard negative samples with batch sampling, following ALBEF~\cite{li2021align}.

\minisection{Image-Text Contrastive Loss (ITC)} ensures that visual and textual embeddings share the same semantic space and the positive (matched) image-text pairs are pulling closer than negative (unmatched) ones. We use two queues $I_q, T_q$ to save the latest visited image and text samples. For each image-text pair $(I, T)$, the softmax-normalized cross-modal similarity is computed as as:
\begin{equation}
  \begin{aligned}
    \label{eqn:itm}
    p_{i2t} (I,T,T_q) &= \frac{exp(sim(I,T)/ \tau)}{\sum_{T' \in T_q} exp(sim(I,T')/\tau)}\\
    p_{t2i} (T,I,I_q) &= \frac{exp(sim(T,I)/\tau)}{\sum_{I' \in I_q} exp(sim(T,I')/\tau)}\\
  \end{aligned}
  ,
\end{equation}
where $\tau$ is a temperature parameter and $sim(\cdot)$ measures cross-modal similarity, which is implemented by the dot product between image and text embeddings.
% Let $y_{i2t}(I),y_{t2i}(T)$ denotes ground truth similarity of image $I$ and text $T$. 
Following ALBEF \cite{li2021align}, we compute ITC loss as:
\begin{equation}
  \begin{aligned}
    \label{eqn:itc}
    \mathcal{L}_{itc}(I,T) &= -log(p_{i2t}(I,T,T_q))-log(p_{t2i}(T,I,I_q)).
  \end{aligned}
\end{equation}

\minisection{Language Modeling Loss (LM)} encourages the model to predict masked words with context information. We randomly mask $15\%$ text tokens and apply the masked language modeling loss as follows:
\begin{equation}
  \begin{aligned}
    \label{eqn:mlm}
    \mathcal{L}_{lm}(I,T) &= H(p_{mask}(I,T),y_{mask}),
  \end{aligned}
\end{equation}
where $y_{mask}$ denotes the masked word to predict and $p_{mask}(I,T)$ is its predicted probability.

The full pre-training objective is the combination of the downstream loss and our constraint on progressive region regression, computed as follows:
\begin{equation}
  \begin{aligned}
    \label{eqn:itm}
    \mathcal{L}_{all}&= \mathcal{L}_{ds} +\mathcal{L}_{reg} \\
  \end{aligned}
 \end{equation}
where $\mathcal{L}_{all}$ is the downstream loss and $\mathcal{L}_{reg}$ denotes the summation of the regression loss in Equation~\ref{eqn:re} for all regions.

\begin{equation}
  \begin{aligned}
    \label{eqn:itm}
    \mathcal{L}_{ds} (I,T) &= \mathcal{L}_{itm} (I,T)+\mathcal{L}_{itc}(I,T)+\mathcal{L}_{lm}(I,T).
  \end{aligned}
\end{equation}

\section{Experiments} \label{sec:experiments}

%In this section, we first introduce our pre-training setup. Then we list experiment results on some cross-modal benchmarks. The ablation studies are carried to incrementally verify our approach and thoroughly investigate the behavior of SCAN.
% We also compare with other methods using fine-grained annotations. Our method improves the performance of cross-modal and localization tasks, showing its effect in both ways. 
We first pre-train our method on a combined dataset of 14M image-text pairs from five datasets: COCO~\cite{lin2014microsoft}, Visual Genome~\cite{krishna2017visual} (excluding
COCO images), Conceptual Captions~\cite{changpinyo2021conceptual}, Conceptual~\cite{changpinyo2021conceptual}, and SBU Captions~\cite{ordonez2011im2text}. The data statistics are described in supplementary. We evaluate the proposed SLAN by comparing it to other state-of-the-art cross model methods on several downstream tasks.  The ablation studies are further conducted to study how each component of our method influences the performance.

\begin{table*}[t]
	\centering
	\small
	\renewcommand{\arraystretch}{1.1}
  \setlength\tabcolsep{1.2mm}

	\begin{tabular}{cc|cccccc|cccccc} \whline{1pt}
	  \multirow{2}{*}{Method} & \multirow{1}{*}{Pre-train} & \multicolumn{6}{c|}{Zero-shot}&\multicolumn{6}{c}{Fine-tune}    \\
% 	  \cline{4-5}
	   &\# Images &\multicolumn{3}{c}{Image $\rightarrow$ Text}&\multicolumn{3}{c|}{Text $\rightarrow$ Image}&\multicolumn{3}{c}{Image $\rightarrow$ Text}&\multicolumn{3}{c}{Text $\rightarrow$ Image} \\ 
	   \whline{0.7pt}
	   &&R@1&R@5&R@10&R@1&R@5&R@10&R@1&R@5&R@10&R@1&R@5&R@10\\
% CLIP&400M& 88.0& 98.7& 99.4& 68.7& 90.6& 95.2& 58.4& 81.5& 88.1& 37.8& 62.4& 72.2\\
ALIGN~\cite{li2021align} &1.8B&88.6& 98.7& 99.7& 75.7& 93.8& 96.8& 95.3& 99.8& 100.0& 84.9& 97.4& 98.6\\
FILIP~\cite{yao2021filip} &300M& 89.8& 99.2& 99.8& 75.0& 93.4& 96.3& 96.6& 100.0& 100.0& 87.1& 97.7& 99.1\\
	  BLIP~\cite{li2022blip} &14M& 94.8& 99.7& 100.0& 84.9& 96.7& 98.3& 96.6& 99.8 &100.0& 87.2& 97.5& 98.8\\
	  BEIT-3~\cite{wang2022image}&21M& 94.9 &99.9& 100.0& 81.5& 95.6& 97.8& 98.0& 100.0& 100.0& \textbf{90.3}& 98.7& 99.5\\
	  Ours&14M&\textbf{96.0}&\textbf{100.0}&\textbf{100.0}& \textbf{86.1}&\textbf{97.0}&\textbf{98.5}&\textbf{98.1}&\textbf{100.0}&\textbf{100.0 }&90.2&\textbf{99.0}&\textbf{99.6}\\
	  \whline{1pt}
	\end{tabular} 
	\caption{\label{tab:ftr}Comparison with state-of-the-art image-text retrieval methods on Flickr30k. We use Recall@k scores as the evaluation metric under zero-shot and fine-tuning settings.}
  \end{table*}
\begin{table*}[t]
	\centering
	\small
	\renewcommand{\arraystretch}{1.1}
  \setlength\tabcolsep{1.2mm}

	\begin{tabular}{cc|cc|cccc|cc|cc} \whline{1pt}
	  \multirow{2}{*}{Method}&\multirow{1}{*}{Pre-training} &\multicolumn{2}{c|}{Retrieval(COCO)} &\multicolumn{4}{c|}{Caption(COCO)}&\multicolumn{2}{c|}{VQA(VQAv2)}&\multicolumn{2}{c}{NLVR(NLVR2)}   \\
	   &Data&I2T R@1&T2I R@1&B@4&M&C&S&test-dev&test-std&dev&test-P
	   \\ 
	   \whline{0.7pt}
	   %&&R@1&R@5&R@10&R@1&R@5&R@10\\
	   Oscar~\cite{li2020oscar} &6.5M&73.5&57.5& 37.4&30.7& 127.8&23.5& 73.6& 73.8& 79.1& 80.3\\
	   VinVL~\cite{zhang2021vinvl} &8.85M&75.4&58.8& 38.5&30.4&130.8&23.4&  76.5& 76.6& 82.6& 83.9\\
	   SimVLM~\cite{wang2021simvlm}&1.8B &-&-& 40.6&33.7& 143.3&25.4&  80.0& 80.3& 84.5& 85.1\\
	   GLIPv2-H~\cite{zhang2022glipv2}&16M & -&-&-&-&131.0&-&74.6 & 74.8&-&-\\
CoCa~\cite{yu2022coca} &4.8B&-&-& 40.9&33.9& 143.6&24.7& 82.3& 82.3& 86.1& 87.0\\
BLIP~\cite{li2022blip} &14M&82.4&65.1& 40.4&-& 136.7&-& 78.2& 78.3& 82.1&82.2\\ 

% BEIT-3& 44.1&147.6&  84.1& 84.0& 91.5& 92.5\\
Ours&14M&\textbf{85.7}&\textbf{69.2}&\textbf{43.7}&\textbf{34.1}& \textbf{144.3}&\textbf{25.6}&\textbf{83.4 }&\textbf{83.5}&\textbf{90.4}&\textbf{91.3}\\
	  \whline{1pt}
	\end{tabular} 
	\caption{\label{tab:od}Comparison on more downstream tasks. For COCO retrieval, I2T and T2I represent image to text and text to image retrieval task, respectively. For COCO image captioning, we report BLEU@4 (B@4), METEOR (M), CIDEr (C), and SPICE (S) on the Karpathy test split. For VQA, we evaluate the vqa-score on VQAv2 test-dev and test-standard (test-std) splits. For NLVR, we report accuracy on NLVR2 development set (dev) and public test set (test-P).}
  \end{table*}
\subsection{Implementation Details}
% We pre-train our model on 8 NVIDIA Tesla V100 GPUs. 
We choose $BERT_{base}$~\cite{kenton2019bert} as our text encoder, which is initialized from HuggingFace~\cite{wolf2020transformers}. 
For the vision encoder, we explore four options: one is the CNN model ResNet50, and three kinds of ViT:
ViT-Base, ViT-Large and ViT-Huge, which are all random initialized. As for the neighbour size for each region adaptor, we use a ratio $r_i$ to denote them: ($NH_i$, $NW_i$) = ($r_i H_i$, $r_i W_i$), where $r_1, r_2, r_3 = 1, 0.5, 0.25$, respectively. 
We pre-train SLAN for 20 epochs. For different options of the vision encoder, the batch size is set to 1280, 960, 640, 640 for ResNet50, Vit-B, Vit-L and Vit-H, respectively.
The AdamW optimize is adopted with the initial learning rate 3e-4,  and the learning rate is linearly decayed to 0. We resize the input images to 224$\times$224.

\subsection{Comparison on The Downstream Tasks}
We compare our approach with state-of-the-art methods on five challenging cross-modal understanding tasks, including image-text retrieval, image captioning, visual question answering, natural language visual reasoning, zero-shot video-text retrieval. Besides, we transfer our method to two localization tasks: object detection and phrase grounding. The default vision encoder is Vit-Huge, if not specified.
\subsubsection{Image-Text Retrieval}
Given an image, the task expects to retrieve the corresponding text from the text gallery through the input image, and vice versa. We evaluate our method on Flickr30k~\cite{plummer2015flickr30k} under zero-shot and fine-tune settings with Karpathy split and the performance is evaluated in terms of Recall@k. The comparative results are shown in Table~\ref{tab:ftr}. 
%We apply our method on COCO and Flickr30k using Karpathy split, listing zero-shot experiment in Table~\ref{tab:ftr}. 
%on less or equal pre-training data.
%We conclude this as the self-localization ability of our method during pre-training. Fine-tuning results on COCO in Table~\ref{fig:cba} also demonstrate the superiority of our method.
Specifically, on the same pre-training setting, SLAN also outperform BLIP~~\cite{li2022blip} by 3.3\% in average recall@1 on COCO. This performance gain explains the efficiency of learning fine-grained alignments from coarse annotations.  
\subsubsection{Image Captioning}
Given an input image, this task generates a sentence description to describe the image in detail. We use COCO Karpathy split to fine-tune and evaluate our method. Our SLAN outperforms most previous methods under this efficient setting, as shown in Table~\ref{tab:od}. 
\begin{table}[tp]
	\centering
	% \footnotesize
	\small
	\setlength\tabcolsep{1.3mm}
	\renewcommand{\arraystretch}{1.3}
	
	\begin{tabular}{c|cccc}
	\whline{1pt}
	   Method & R@1 $\uparrow$ & R@5 $\uparrow$ & R@10 $\uparrow$ & MdR $\downarrow$ \\ \hline
	   ClipBERT~\cite{lei2021less}& 22.0& 46.8& 59.9& 6\\
	   VideoCLIP~\cite{xu2021videoclip}& 30.9& 55.4 &66.8& -\\
	   FiT\dag~\cite{bain2021frozen}&43.3& 65.6& 74.7& 2\\
	   BLIP\dag~\cite{li2022blip} &43.3& 65.6& 74.7& 2 \\
	   Ours\dag &\textbf{46.8}&\textbf{70.5}& \textbf{83.6}& \textbf{1.5}\\
	\whline{1pt}
	\end{tabular}
	\caption{\label{tab:vr}Comparisons with state-of-the-art methods for text-video retrieval on the 1k test split of the MSRVTT~\cite{xu2016msr-vtt} dataset. \dag denotes the zero-shot settings, while others are fine-tuned ones.
	}
  \end{table}

  \begin{table*}[t]
	\centering
	\small
	\renewcommand{\arraystretch}{1.1}
  \setlength\tabcolsep{1.2mm}

	\begin{tabular}{ccc|cc|cc|p{11mm}p{11mm}p{11mm}} \whline{1pt}
	  \multirow{2}{*}{Method} & \multirow{2}{*}{Backbone}&\multirow{2}{*}{Params(M)}&\multicolumn{2}{c|}{Pretrain Data(M)}& \multicolumn{2}{c|}{Object Detection(COCO)}&\multicolumn{3}{c}{Phrase Grounding(Flickr30k)}
% 	  \\\cmidrule{4-5} \cmidrule{7-7}
\\
	   &&&Image-Text&Region-Word&Zero-shot&Fine-tune&R@1&R@5&R@10
	   \\ 
	   \whline{0.7pt}
	   DETR~\cite{carion2020end}&ResNet50&42&0&0&-&42.0&-&-&-\\
	   MDETR~\cite{kamath2021mdetr}&ResNet101&185&0&0.2&-&-&84.3&93.9&95.8\\
	   GLIP~\cite{li2022grounded} &Swin-Large&430&24&3& 49.8& 60.8& 87.1& 96.9&98.1\\
	   GLIPv2~\cite{zhang2022glipv2} &Swin-Huge&870&16&3& -& 60.2& 87.7&97.3&98.5\\\hline
	   Ours &ResNet50&322&14&0& 46.9& 59.2& 86.8&96.6&97.4\\
	   Ours &Vit-Base& 383&14&0&47& 59.6& 87.4&96.9&98.2\\
	   Ours&Vit-Large&601&14&0& 48.5& 60.5& 89.1&98.0&98.9\\
	   Ours&Vit-Huge&929&14&0& \textbf{50.1}& \textbf{63.5}& \textbf{90.6}&\textbf{98.6}&\textbf{99.3}\\
	  \whline{1pt}
	\end{tabular} 
	\caption{\label{tab:det}Comparison on two localization tasks: Object Detection on COCO and Phrase Grounding on Flickr30k. The pre-training data covers: image-text pairs and word-specific region annotations. we evaluate zero-shot and fine-tune  settings on object detection. We use recall@k scores to evaluate phrase grounding task.
	}
  \end{table*}

\begin{table}[t]
	\centering
	\small
	\renewcommand{\arraystretch}{1.1}
  \setlength\tabcolsep{1.2mm}

	\begin{tabular}{cccccccc} \whline{1pt}
	  \multirow{1}{*}{Trainable} & \multirow{1}{*}{Adaptor}& &\multicolumn{2}{c}{COCO}&&\multicolumn{2}{c}{Flickr30k}
% 	  \\\cmidrule{4-5} \cmidrule{7-8}
\\
	   Detector&Number&&TR@1&IR@1&&TR@1&IR@1
	   \\ 
	   \whline{0.7pt}
	   \ding{56}&0&&68.5&53.5&&85.0&74.1\\
	   %\ding{52}&0&&69.3&54.1&&86.4&75.9\\
	   %\ding{52}&1&&70.6&57.6&&88.7&77.5\\
	   %\ding{52}&2&&71.0&57.8&&89.2&77.6\\
	   %\ding{52}&3&&72.0&58.1&&90.4&78.9\\
	   \ding{52}&0&&69.1&53.8&&86.7&76.2\\
	   \ding{52}&1&&70.0&57.2&&88.3&77.4\\
	   \ding{52}&2&&70.8&57.5&&88.7&78.1\\
	   \ding{52}&3&&\textbf{72.1}&\textbf{58.3}&&\textbf{90.3}&\textbf{78.9}\\
	  \whline{1pt}
	\end{tabular} 
	\caption{\label{tab:ba}Effect of training detection module and region adaptor. \ding{56} in the first coloum denotes applying a frozen detection module and no self-locator. TR@1 and IR@1 denote recall@1 of image to text and text to image retrieval. To evaluate the effect of the self-locator against a frozen detection module, we load the pre-trained weight from COCO Detection to compare with our method for line 1. The remaining experiments train from scratch. We use Vit-Base as the vision encoder.}
  \end{table}
  
\begin{table}[t]
	\centering
	\small
	\renewcommand{\arraystretch}{1.1}
  \setlength\tabcolsep{1.2mm}

	\begin{tabular}{cccccccc} \whline{1pt}
	  \multirow{2}{*}{Top K} & \multirow{2}{*}{Threshold}& &\multicolumn{2}{c}{COCO}&&\multicolumn{2}{c}{Flickr30k}
\\
	   &&&TR@1&IR@1&&TR@1&IR@1
	   \\ 
	   \whline{0.7pt}
	   -&-&&69.4&54.1&&85.9&74.7\\
	   10&-&&70.6&56.8&&87.5&77.3\\
	   10&0.3&&71.2&57.6&&89.1&78.2\\
	   10&0.5&&\textbf{72.1}&\textbf{58.3}&&\textbf{90.3}&\textbf{78.9}\\
	  \whline{1pt}
	\end{tabular} 
	\caption{\label{tab:bf}Different settings of the region filter.}
  \end{table}

\subsubsection{Visual Question Answering}
Visual Question Answering (VQA)~\cite{antol2015vqa} requires the model to predict an answer from an image and a question. We follow ~\cite{li2022blip} and treat VQA as an open-ended question-generation task. We fuse the image embedding with the question embedding and send them to the question decoder to get the result. As shown in Table~\ref{tab:od}, SLAN outperform other method by at least 1.1\% on VQAv2 test-dev and test-std with less or equal pre-training data.

\subsubsection{Natural Language Visual Reasoning}
Natural Language Visual Reasoning (NLVR2)~\cite{suhr2018corpus} measures whether a sentence describes a pair of images. We extract the image and text embeddings from the image-text input, which are fused by a cross-attention layer. We use a binary classification module to predict their relations. As shown in Table~\ref{tab:od}, our results surpass others by a large margin, showing the importance of learning fine-grained cross-modal alignments. 

\subsubsection{Zero-shot Video-Text Retrieval}
Besides the image-text tasks mentioned above, our method can generalize to the video-text retrieval task. We randomly select $m$ frames of the video input and concatenate them to get an image-text sequence, then we feed them directly into our image-text retrieval model. The performance in Table~\ref{tab:vr} is comparable to others, demonstrating the cross-modal knowledge learned in our method is semantic-rich.

\subsubsection{Localization Tasks}
 We conduct two localization tasks: object detection on COCO, and phrase grounding on Flickr30k. 
For the text input on the object detection task, we use a prompt composed of concatenated labels of COCO (e.g. detect:  person, bicycle, car, ... , toothbrush). We adopt the output from the last layer of the Table~\ref{tab:det} has exciting results about SLAN on localization tasks.  
% This advancement is achieved by learning detailed region-to-word alignments during pre-training with our SLAN. 
For example, in the task of object detection with Vit-Base as the backbone, SLAN achieves comparable results to GLIP using a larger backbone and 3M gold data, i.e., Swin-Large, only slightly worse on object detection task, but still slightly better on grounding task. When applying a larger backbone Vit-Huge, our method significantly outperforms all comparison methods.

\subsection{Ablation Study}
% In this section, we analyse some inner factor of our method.
\subsubsection{Effect of Self-locator
}

\minisection{Importance of learnable detection module.} As shown in the first row of Table~\ref{tab:ba}, we replace our self-locator with a frozen detector pre-trained on COCO Detection, and the second row is the result of our learnable detector. We do not load the detector with pre-trained weights, but only fine-tune on the downstream task datasets. Our method improves on average about 0.5\% and 2\% on COCO and Flickr30k's image-to-text
and text-to-image retrieval tasks, respectively. 
% We ablate on all components helping the framework produce adapted region proposals in Table~\ref{tab:ba}. To verify their effect, we do not load pre-trained weights of 14M data. The first two records show the performance gain brought by fine-tuning the detection module without the supervised signal from the region adaptor. The detection module is updated with gradients from the RoIAlign operation. The next three records compare the number of region adaptor layer $L$. Our region adaptor improves the performance on retrieval tasks. Results also demonstrate the effect of increasing the number of region adaptor, showing that splitting region adaptation into more steps eases its learning. 

\minisection{Number of region adaptors for region regression.} The region adaptor performs progressive regression on the regions output by the detector to provide more accurate region localization for cross-modal understanding tasks. As shown in Table~\ref{tab:ba}, as the number of region adaptors increases from 0 to 3, the retrieval performance can be significantly improved by an average of more than 3\%.

\minisection{Region filter for saliency prediction.}
Table~\ref{tab:bf} illustrates how the region filter affects the performance on COCO and Flickr30k retrieval tasks. Learnable detector is trained from scratch and the number of region adaptors is set to 3. The first and second rows show that when the regions are sorted by saliency score and selected by a certain number, we can achieve an performance gain of ~2\% on each dataset. When using the saliency score threshold, our region filter is able to remove redundant regions that negatively affect cross-modal adaptation with higher performance.
% We ablate on the threshold of the region filter, which removes some irrelevant regions for cross-modal adaptation. In Table~\ref{tab:bf}, we show that two strategies for filtering region collaborate to provide adapted region proposals for cross-modal alignments. Each experiment adopts a trainable detection module and three layers of region adaptor. We train them from scratch for fair initializations.

\subsubsection{Computational Cost}
Table~\ref{tab:com} illustrates the computational cost of our method and other state-of-the-art methods. Our method has the smallest amount of parameters and FLOPS because in this experiments our vision backbone is a relatively lightweight ResNet50. However, our retrieval performance significantly outperforms other methods. As far as our concerned, our method is efficient and high-performance.
% In Table~\ref{tab:com}, we compare some VLP methods on computational cost. We adopt the image-to-text retrieval tasks on COCO as an example. Our method with ResNet50 as the vision backbone outperforms all other methods. It also has fewer parameters and FLOPS, showing the contribution of learning detailed alignments between cross-modal data.
\begin{table}[t]
	\centering
	\small
	\renewcommand{\arraystretch}{1.1}
  \setlength\tabcolsep{1.2mm}

	\begin{tabular}{cccccc} \whline{1pt}
	  \multirow{2}{*}{Method}&\multirow{2}{*}{Backbone} & \multirow{2}{*}{Params(M)}& \multirow{2}{*}{FLOPS(G)}&\multicolumn{2}{c}{COCO}
% 	  \\\cmidrule{4-5} \cmidrule{7-8}
\\
	   &&&&TR@1&IR@1
	   \\ 
	   \whline{0.7pt}
	   BLIP&Vit-Base&370&558&81.9&64.3\\
	   BLIP&Vit-Large&810&1594&82.4&65.1\\
	   Coca&Vit-Huge&2100&4103&83.0&65.5\\
	   Beit-3&Vit-Huge&1900&-& 84.8&67.2\\
	   Ours&ResNet50&322&324&\textbf{85.1}&\textbf{68.9}\\
	  \whline{1pt}
	\end{tabular} 
		\caption{\label{tab:com}Comparison of model parameter and FLOPS on cross-modal retrieval task. The FLOPS is calculated with the image of 384x384 resolution. The Backbone denotes the vision encoder.}
  \end{table}
\subsection{Visualization Analysis}
\subsubsection{Text-guided Region Adaptation}
% The coordinates of regions are updated under the guidance of text embedding. Given the same region proposals from the detection module, we visualize two adaptation results with different text inputs. 
As shown in Figure~\ref{fig:tg}, our region adaptor produces text-specific results with relatively high confidence.  When we change the detailed description of the sentence, e.g., ``a man in a red coat" to ``a man in black pants", the interesting phenomenon is that the attention regions of our self-locator are also shifted accordingly with relatively high confidence.

\subsubsection{Region adaptation during training phases}
SLAN learns to localize accurate regions given input sentences. As illustrated in Figure~\ref{fig:pha}, as training progresses, the regions of interest outputed by our model gradually become accurate. This demonstrates that our model can be trained without grounding annotation and can gradually learn fine-grained image-text alignment information.

\newcommand{\addtg}[1]{\includegraphics[width=0.47\linewidth]{fig/tg/#1.jpg}}
\begin{figure}
	\centering
	\small
	\renewcommand{\arraystretch}{1.1}
  \setlength\tabcolsep{1.2mm}
	
\begin{tabular}{cc}
\addtg{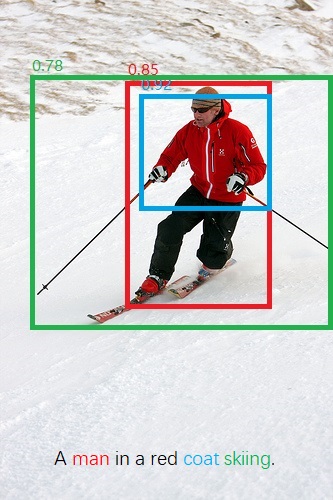}&\addtg{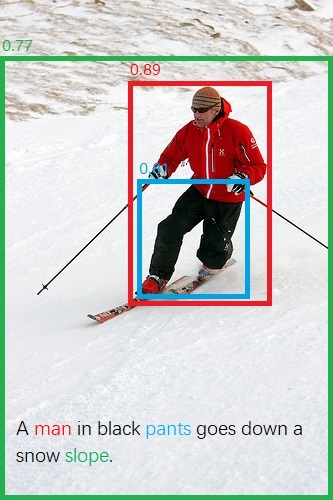}\\
    
    \end{tabular}
    \caption{Illustration of text-specific region adaptation. We colorize three words per sentence and use the corresponding colors to mark the regions with the highest matching scores. This denotes the interpretable region adaptation of our method, which brings fine-grained cross-modal feature fusion for downstream tasks.}
	\label{fig:tg}
\end{figure}

\newcommand{\addtrb}[1]{\includegraphics[width=0.31\linewidth]{fig/trb/#1.jpg}}
\begin{figure}
	\centering
	\small
	\renewcommand{\arraystretch}{1.1}
  \setlength\tabcolsep{1.2mm}
	
\begin{tabular}{ccc}
\addtrb{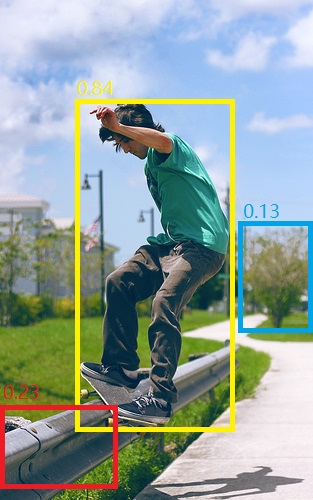}&\addtrb{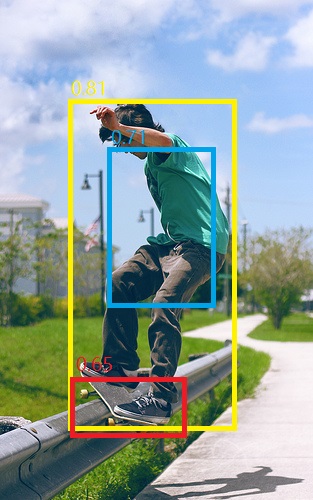}&\addtrb{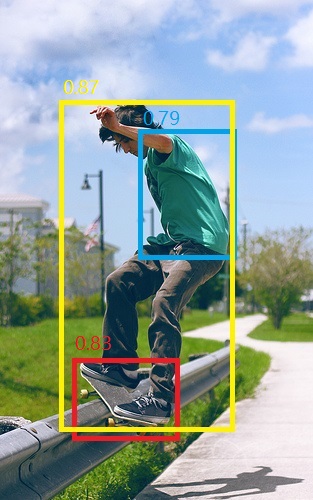}\\
    0&1/2&1\\
    
    \end{tabular}
    \caption{Visualization of region proposal during pre-training. The numbers under images denote the relative training duration, where $1$ represents the whole pre-training procedure. The referenced text is "A \textcolor{yellow}{man} in a green \textcolor{blue}{shirt} is doing the trick on a \textcolor{red}{skateboard}."}
	\label{fig:pha}
\end{figure}
\begin{figure}[!t]
    \centering
    \includegraphics[scale=1.55]{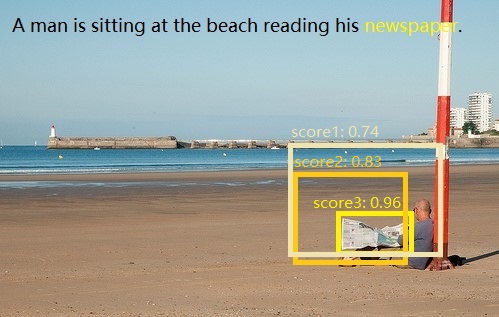}
	\caption{ \label{fig:cba}Coarse-to-fine process of the region adaptation. We also list the matching score between the regions and counterpart words. 
	}
\end{figure}

\subsubsection{Coarse-to-fine Region Adaptation}
To verify the calibration effect of region adaptation, we visualize an image with its text in Figure~\ref{fig:cba}. Model locates more accurate regions of interest with higher similarity scores after three levels of region adaptor. This shows that our self-locator hierarchically refines the related regions of provided words.

% \subsubsection{Openset Detection}

\section{Conclusions and Future Work}
In this paper, we present the Self-Locator Aided Network (SLAN), which incorporates the self-locator to adapt proposed regions for cross-modal alignments without extra grounding (region-to-word) annotations. Experimental results show that our method achieves a strong performance on many cross-modal understanding tasks. Besides, the self-locator can also be transferred to some localization tasks, and demonstrates its potential on them. We hope SLAN provides a new direction for learning fine-grained cross-modal alignments without relying on expensive grounding annotations.

\clearpage
{\small
    \bibliographystyle{ieee_fullname}
    \bibliography{mm}
}
\clearpage
\newpage
\appendix
\section{Appendices}

\subsection{Details of Pre-training Dataset}
We list the statistic of each dataset used for pre-training. The overall pre-training dataset has 14M image-text pairs, without any fine-grained region-to-word annotations. 
\begin{table}[t]
	\centering
	\small
	\renewcommand{\arraystretch}{1.1}
  \setlength\tabcolsep{1.2mm}

	\begin{tabular}{cccccccc} \whline{1pt}
	   & \#Image&\#Text
% 	  \\\cmidrule{4-5} \cmidrule{7-8}
\\
	   \whline{0.7pt}
	   COCO&113K&567K\\
	   Visual Genome & 100K & 769K\\
	   SBU captions & 860K & 860K\\
	   Conceptual Captions & 3M & 3M\\
	   Conceptual &10M&10M \\

	  \whline{1pt}
	\end{tabular} 
	\caption{\label{tab:da}The number of images and texts in our pre-training datasets.}
  \end{table}
  
\subsection{Load Detection Weights for Pre-training}

\begin{table}[t]
	\centering
	\small
	\renewcommand{\arraystretch}{1.1}
  \setlength\tabcolsep{1.2mm}

	\begin{tabular}{cccccccc} \whline{1pt}
	  \multirow{1}{*}{Pre-trained} & &\multicolumn{2}{c}{COCO}&&\multicolumn{2}{c}{Flickr30k}
% 	  \\\cmidrule{4-5} \cmidrule{7-8}
\\
	   Detector&&TR@1&IR@1&&TR@1&IR@1
	   \\ 
	   \whline{0.7pt}
	   \ding{56}&&85.1&68.9&&97.4&88.2\\
	   \ding{52}&&85.6&67.7&&97.6&88.4\\
	  \whline{1pt}
	\end{tabular} 
	\caption{\label{tab:pre}Comparison on loading the pre-trained detection module. We set the vision encoder as ResNet50.}
  \end{table}

We pre-train our SLAN from scratch in default. Besides, we also attempt to load the pre-trained weights on COCO Detection~\cite{lin2014microsoft} for the detection module (DETR~\cite{carion2020end}), which provides more prior knowledges to learn the overall architecture. As shown in Table~\ref{tab:pre}, loading pre-trained detection module brings 0.5\% and 0.8\% on recall@1 of COCO image-text retrieval, while training from scratch also performs relatively well. This also demonstrates that our SLAN can learn rich cross-modal alignments without any pre-defined information.

\begin{table}[t]
	\centering
	\small
	\renewcommand{\arraystretch}{1.1}
  \setlength\tabcolsep{1.2mm}

	\begin{tabular}{cccccccc} \whline{1pt}
	  \multirow{1}{*}{Feature} & &\multicolumn{2}{c}{COCO}&&\multicolumn{2}{c}{Flickr30k}
% 	  \\\cmidrule{4-5} \cmidrule{7-8}
\\
	   Source&&TR@1&IR@1&&TR@1&IR@1
	   \\ 
	   \whline{0.7pt}
	   Encoder&&82.4&66.7&&96.2&87.5\\
	   Decoder&&85.1&68.9&&97.4&88.2\\
	  \whline{1pt}
	\end{tabular} 
	\caption{\label{tab:msf}Different source of multi-scale vision feature on SLAN.}
  \end{table}

\subsection{Multi-scale Vision Feature}
We ablate with two kinds of multi-scale vision feature used in the region adaptor. The default setting in the paper is to introduce a vision decoder to provide rich visual semantic knowledge. To prove its necessity, we use the vision encoder for generating multi-scale vision feature and list results in Table~\ref{tab:msf}. The experiment with decoder vision feature outperforms the other one with more than 2\%  on recall@1 of COCO image-text retrieval.
\newcommand{\addrt}[1]{\includegraphics[width=0.18\linewidth]{fig/retrieval/#1.png}}

\subsection{Visualization of Region Filter}

\newcommand{\addft}[1]{\includegraphics[width=0.5\linewidth]{fig/filter/#1.png}}
\begin{figure}
	\centering
	\small
	\renewcommand{\arraystretch}{1.1}
  \setlength\tabcolsep{1.2mm}
\begin{tabular}{cc}
\addft{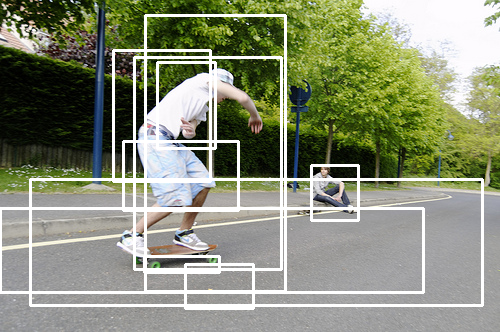}&\addft{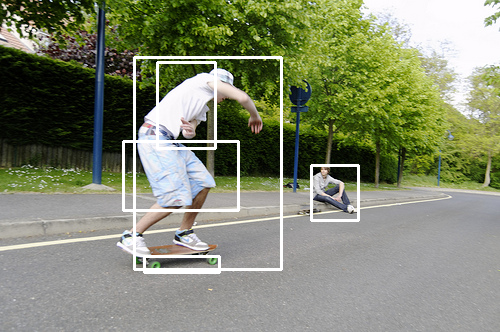}\\
    Before Region Filter& After Region Filter\\
    \end{tabular}
    \caption{
    Illustration of the image before and after our region filter. The caption is ``A guy wearing shorts and a white t-shirt is skateboarding down the road, while someone sits and watches him from the curb".
    }
	\label{fig:ft}
\end{figure}
As shown in Figure~\ref{fig:ft}, our region filter selects salient regions for further adaptation with words. It effectively focuses on informative regions.

\begin{figure}
	\centering
	\small
	\renewcommand{\arraystretch}{1.1}
  \setlength\tabcolsep{0.4mm}
\begin{tabular}{cccccc}
\rotatebox{90}{~~~BLIP}&\addrt{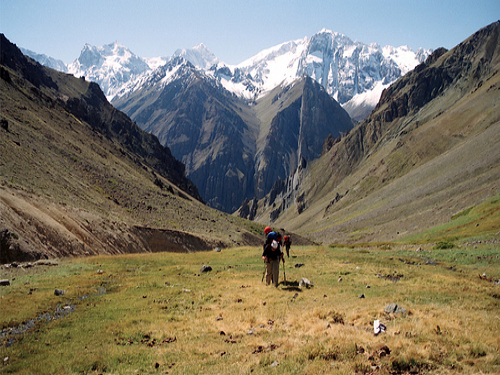}&\addrt{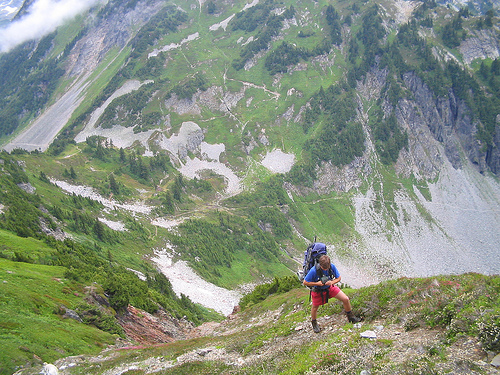}&\addrt{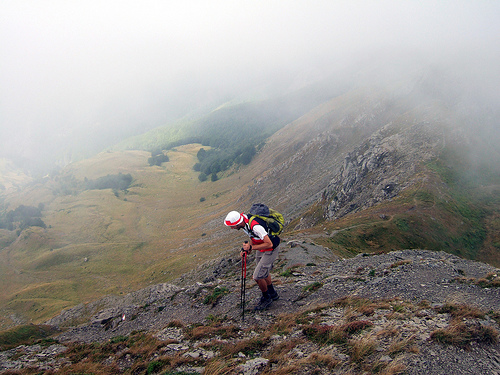}&\addrt{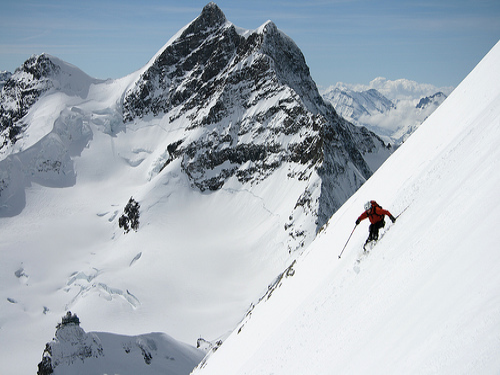}&\addrt{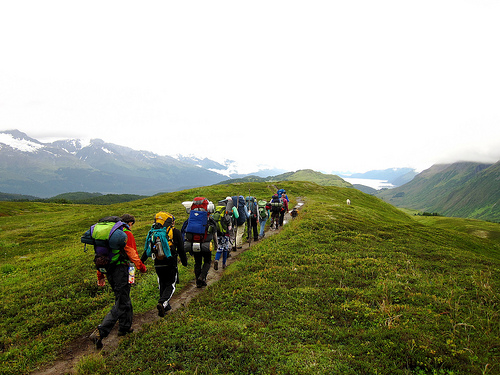}\\
\rotatebox{90}{~~~Ours}&\addrt{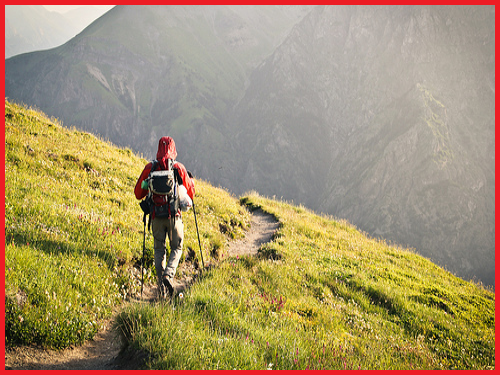}&\addrt{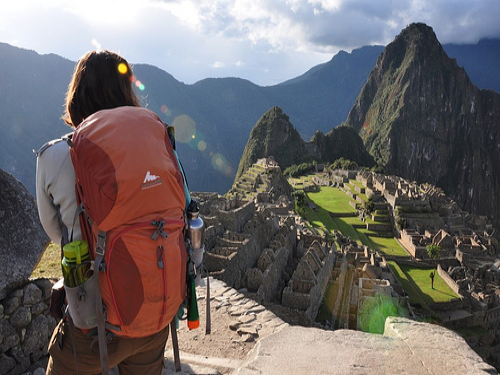}&\addrt{o3}&\addrt{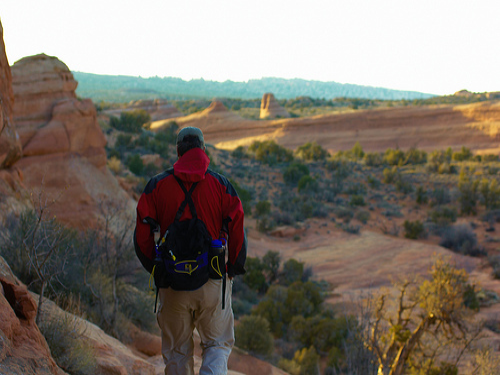}&\addrt{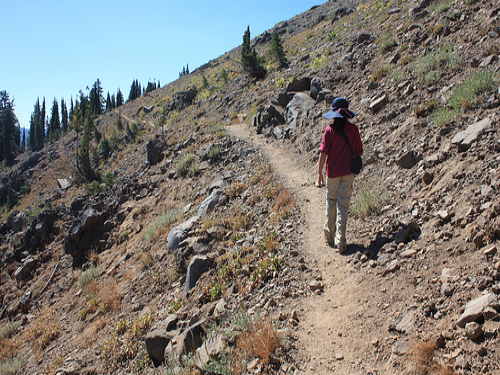}\\
   &\multicolumn{5}{c}{Text: A lonely man heads off on a mountain trail wearing a}\\
   &\multicolumn{5}{c}{bright red jacket.}\\
    \end{tabular}
    \caption{
    Text-to-image retrieval on Flickr30k. We list top-5 retrieved images given query text for two methods, and highlight the ground truth with red box.
    }
	\label{fig:retrieval}
\end{figure}
\newcommand{\addgr}[1]{\includegraphics[width=0.5\linewidth]{fig/gr/#1.png}}
\begin{figure}
	\centering
	\small
	\renewcommand{\arraystretch}{1.1}
  \setlength\tabcolsep{0.4mm}
\begin{tabular}{cc}
\addgr{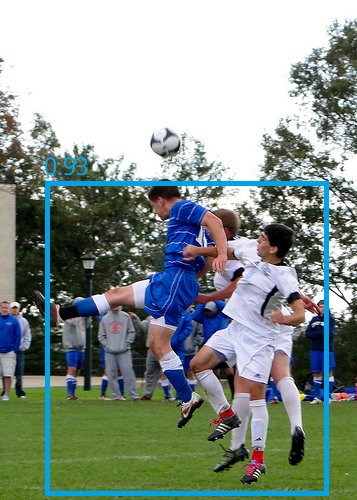}&\addgr{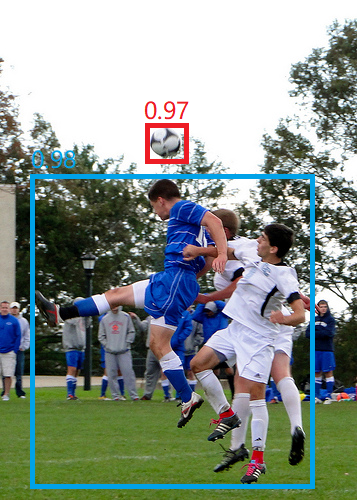}\\
MDETR&Ours\\
    \end{tabular}
    \caption{
    We compare our method with MDETR on phrase grounding in Flickr30k. The query text is: A group of \textcolor{blue}{people} are playing \textcolor{red}{soccer} on the field.
    }
	\label{fig:gr}
\end{figure}

\subsection{Visualization of image-text retrieval and Phrase Grounding}
We list the retrieval results of our SLAN and BLIP~\cite{li2022blip} for comparison. SLAN can capture more fine-grained relation between the query text and images, (\eg{red jacket in retrieved images}). This leads to more accurate retrieval performance and our method successfully marks the ground truth as the top-1 retrieval result. 

To show the detail-aware localization ability of SLAN, we also compare with MDETR~\cite{kamath2021mdetr} in Figure~\ref{fig:gr}. Both methods recognize the people in the image, while our SLAN also locates the soccer with relatively high confidence.

\end{document}